%% file: main.tex
\documentclass{article}

    \usepackage[preprint]{neurips_data_2024}

\usepackage[utf8]{inputenc} 
\usepackage[T1]{fontenc}    
\usepackage{hyperref}       
\usepackage{url}            
\usepackage{booktabs}       
\usepackage{amsfonts}       
\usepackage{nicefrac}       
\usepackage{microtype}      
\usepackage{xcolor}         

\usepackage{graphicx}
\usepackage{appendix}

\input{macros}

\title{Psychometric Alignment: Capturing Human Knowledge Distributions via Language Models}

\author{
  Joy He-Yueya$^1$\thanks{Corresponding author: \texttt{\href{mailto:heyueya@cs.stanford.edu}{\textcolor{black}{heyueya@cs.stanford.edu}}}} 
\hspace{1cm} Wanjing Anya Ma$^2$
\hspace{1cm} Kanishk Gandhi$^1$
\\
\textbf{Benjamin W.~Domingue}$^2$
\hspace{1cm} \textbf{Emma Brunskill}$^1$
\hspace{1cm} \textbf{Noah D.~Goodman}$^{1,3}$ \\
Departments of Computer Science$^1$, Education$^2$, and Psychology$^3$, Stanford University
}

\begin{document}

\maketitle

\begin{abstract}
Language models (LMs) are increasingly used to simulate human-like responses in scenarios where accurately mimicking a population's behavior can guide decision-making, such as in developing educational materials and designing public policies. The objective of these simulations is for LMs to capture the variations in human responses, rather than merely providing the expected correct answers. Prior work has shown that LMs often generate unrealistically accurate responses, but there are no established metrics to quantify how closely the knowledge distribution of LMs aligns with that of humans. To address this, we introduce ``psychometric alignment,'' a metric that measures the extent to which LMs reflect human knowledge distribution. Assessing this alignment involves collecting responses from both LMs and humans to the same set of test items and using Item Response Theory to analyze the differences in item functioning between the groups. We demonstrate that our metric can capture important variations in populations that traditional metrics, like differences in accuracy, fail to capture. We apply this metric to assess existing LMs for their alignment with human knowledge distributions across three real-world domains. We find significant misalignment between LMs and human populations, though using persona-based prompts can improve alignment. Interestingly, smaller LMs tend to achieve greater psychometric alignment than larger LMs. Further, training LMs on human response data from the target distribution enhances their psychometric alignment on unseen test items, but the effectiveness of such training varies across domains. \footnote{Code and data available here: \url{https://github.com/joyheyueya/psychometric-alignment}}
\end{abstract}

\input{sections/1_introduction}
\input{sections/2_related}
\input{sections/3_psychometric_alignment}

\input{sections/datasets}
\input{sections/4_prompting}
\input{sections/finetuning}
\input{sections/conclusion}
\section{Acknowledgments}
We would like to thank Dongwei Jiang, Simon Woodhead, Ryan Louie, Allen Nie, Scott L.~Fleming, Ben Prystawski, Yann Hicke, Alex Nam, Aishwarya Mandyam, and Jason D.~Yeatman for their feedback and support. This work was supported by the Junglee Corporation Stanford Graduate Fellowship, a Stanford Hoffman-Yee grant, and the Openai Researcher Access Program. This work was supported by the Stanford Interdisciplinary Graduate Fellowship to WAM.

\bibliography{main}
\bibliographystyle{plainnat}
\newpage
\appendix

\input{sections/appendix}

\end{document}

%% file: macros.tex
\newcommand{\eedi}{\textsc{Eedi}}
\newcommand{\duolingo}{\textsc{Duolingo}}
\newcommand{\wordbank}{\textsc{WordBank}}
\newcommand{\persona}{$\textsc{PERSONA}$}
\newcommand{\personacot}{$\textsc{PERSONA-COT}$}
\newcommand{\personacots}{$\textsc{PERSONA-COT-S}$}

\def\Snospace~{\S{}}

\definecolor{coco1}{HTML}{D9E4EC}
\definecolor{coco2}{HTML}{B7CFDC}
\definecolor{coco3}{HTML}{6AABD2}
\definecolor{coco4}{HTML}{385E72}
\hypersetup{
    colorlinks=true,
    linkcolor=coco3,
    filecolor=coco4,      
    urlcolor=coco3,
    citecolor=coco3,
}

\usepackage{subcaption, floatrow}
\usepackage{multirow}

%% file: sections/1_introduction.tex
\section{Introduction}
The ability of language models (LMs) to mimic human behaviors has been used to replicate results from social science experiments and public opinion surveys \citep{argyle2023out, aher2023using, horton2023large}, and opens up exciting possibilities in areas such as education \citep{markel2023gpteach, he2024evaluating}, marketing \citep{brand2023using}, and product design \citep{park2022social, park2023generative}. In these applications, LMs have been used to represent human populations and respond to questions in various domains. Unlike typical benchmarks that ta standard ideal or common response, the objective in these settings is for LMs to reflect the distribution of responses and outcomes observed in human populations. For instance, when simulating interactions between a novice student and a teacher, we expect that students with different levels of knowledge will have varying probabilities of producing correct answers. However, prior work has demonstrated that LM-generated responses can sometimes be unrealistically advanced \citep{aher2023using, chuang2023simulating}. It is therefore important to assess the extent to which LMs capture the human population distribution of knowledge or capabilities and develop methods to align LMs with human distributions. If LMs could effectively mimic human distributions of knowledge, it opens up exciting opportunities for LMs to understand and support human learning. 

Measuring the alignment between the knowledge distribution of LMs and that of a human population poses a challenge, as we cannot directly observe the cognitive processes of either group. One intuitive approach is to assess the knowledge of both LMs and humans on a set of test items (questions) and compare the accuracy/score distributions of the two groups. However, simply comparing scores can be misleading for evaluating the similarity between two populations and understanding the impact of specific test items because it fails to capture the distribution of knowledge across individual test items (see Section \ref{sec:the_importance_of_psychometric_alignment}). To address this, we introduce psychometric alignment, an evaluation metric that measures the extent to which LMs capture the knowledge distribution of a human population. Assessing this alignment involves collecting responses from both LMs and humans to a set of test items and analyzing the differences in item functioning across groups. In particular, we use Item Response Theory \citep{lord2012applications} to estimate the item parameters, such as difficulty, for each group and compute the Pearson correlation between these parameters to quantify the extent to which LMs reflect human knowledge distributions. Through psychometric simulations, we demonstrate that this metric is robust, sensitive, and stable in identifying (mis)alignment.

Using this metric, we benchmark the ability of existing LMs to capture human knowledge distributions across three real-world domains: first-language acquisition (\wordbank{} \citep{frank2017wordbank}), second-language learning (\duolingo{} \citep{settles2018second}), and mathematics (\eedi{}). \eedi{} is a new dataset that is built on the NeurIPS 2020 Education
Challenge dataset \citep{wang2020instructions} and contains responses from 2287 students aged 11-12 to 573 math multiple-choice questions. We find a substantial misalignment between the knowledge distribution of an ensemble of LMs that vary in size and capability and that of humans in the mathematics domain.

We then explore prompting and training methods aimed at enhancing the psychometric alignment between LMs and humans in various domains. In particular, we show that creating a set of LM instances using persona-based prompting leads to stronger psychometric alignment. However, this method falls far short of ceiling performance (i.e., the alignment between human subgroups from the same population), and its effectiveness varies significantly across domains and LMs. Interestingly, smaller LMs tend to achieve better psychometric alignment than larger LMs. This suggests that increasing model size/training data, that is intended to make LMs more capable at instruction following and various tasks, may reduce the capacity of LMs to simulate human behaviors accurately. Moreover, fine-tuning LMs on data from the target human distribution leads to further improvement in psychometric alignment on unseen test items, but its effectiveness varies across domains.

%% file: sections/2_related.tex
\section{Related work}

\textbf{Using LMs to simulate humans} \quad
Our work is closely related to recent work on using LMs to simulate human behaviors. For instance, researchers have used LMs to replicate results from social science experiments and public opinion surveys \citep{argyle2023out, aher2023using, horton2023large}. The ability of LMs to mimic human behaviors offers exciting opportunities in areas such as education \citep{markel2023gpteach, he2024evaluating, jin2024teach, zelikman2023generating, lu2024generative,  shaikh2023rehearsal, liu2023improving, xu2023leveraging}, marketing \citep{brand2023using, li2023large}, and product design \citep{park2022social, park2023generative}. Evaluations of such simulations have typically been limited to replicating well-established results from prior studies involing real humans, asking experts to assess believability, or comparing summary statistics such as accuracies on various tests. These methods often overlook or fail to assess the alignment between the knowledge distributions of LMs and the target human populations (see Section \ref{sec:the_importance_of_psychometric_alignment}). Notably, several studies have proposed metrics to measure the alignment between two population distributions. \cite{santurkar2023whose} have developed a metric to measure the alignment of LM opinions with different demographic groups over common topics in public opinion surveys. \cite{safdari2023personality,pellert2023ai} have explored whether LMs can simulate non-cognitive human traits such as personalities.

Our work is also related to research on whether LMs can learn representations of concepts that are aligned with humans, as explored in the field of representational alignment (see \cite{sucholutsky2023getting} for a survey). The capabilities of LMs may be fundamentally different from human capabilities even though they may achieve similar overall accuracy scores on certain benchmarks \citep{anwar2024foundational}. For instance, GPT-4's accuracy in a counting task drops significantly when the correct answer is a less likely number (e.g., 29) compared to more likely numbers (e.g., 100), unlike human performance which tends to remain consistent \citep{mccoy2023embers}. Thus, there is a need to develop more nuanced evaluation metrics that capture the distribution of capabilities across a set of tasks or test items. 

\textbf{Predicting psychometrics} \quad
\cite{lalor2019learning} train an ensemble of DNN models with varying amounts of training data and label corruption to generate responses to linguistic reasoning tasks and then apply Item Response Theory models to estimate the task difficulties from the generated responses. \cite{laverghetta2021can} have considered predicting the psychometric properties of test items (e.g., difficulties) by creating an ensemble of different transformer-based and LSTM-based LMs that vary in the number of trainable parameters and the amount of finetuning data to simulate a human population.

%% file: sections/3_psychometric_alignment.tex
\section{Measuring psychometric alignment}
We briefly review Item Response Theory and introduce our metric for quantifying how well LMs align with a human population distribution of knowledge across a set of test items.

\subsection{Item Response Theory}
Consider a scenario where a group of individuals answers a series of test items. Each response from a person reflects an interaction between their ``ability'' (knowledge or capabilities) and various attributes of the test item such as its difficulty. To assess the abilities of individuals, a simple approach is to count the number of correct answers. However, this method fails to account for variations in item difficulty; some items might test more complex concepts. To address these subtleties, Item Response Theory (IRT) \citep{lord2012applications} offers a psychometric framework widely used in educational assessments and psychological measurements to analyze both the abilities of the individuals and the characteristics of the test items simultaneously. Among various IRT models, we review the simplest one-parameter logistic model (1PL), also called the Rasch model \citep{rasch1960studies}. The 1PL model assumes that the probability of a correct response to an item is determined by the difference between the person's ability $\theta_i$ and the item's difficulty $b_j$, shown in Eq.~\ref{eq:1pl_irt}. 
\begin{equation}
    p(X_{i,j}=1|\theta_i,b_j) = \frac{1}{1 + e^{-(\theta_i - b_j)}}
\label{eq:1pl_irt}
\end{equation}

One of the key features of IRT models is related to the assumption of parameter invariance \citep{rupp2006understanding}, meaning that item and person parameters remain stable even when different groups generate responses under varying measurement conditions. This is potentially a strong assumption. For example, consider administering a math test to two different groups: native English speakers (Group 1) and English language learners (Group 2).  There are scenarios where responses may depend on group membership in a way not captured by Eq.~\ref{eq:1pl_irt}. However, by comparing the item difficulty parameters between these groups, we can evaluate whether parameter invariance holds. If the parameters are highly correlated and the differences in item difficulties between the groups are minor, we know that the parameters are invariant across groups. This indicates that the test items function similarly for both groups, ensuring that the test is not biased against any group \citep{camilli2006test, ma2023roar}. Conversely, significant differences in item parameters indicate a lack of invariance, prompting further analysis through psychometric methods to identify items with differential item functioning (DIF) and exclude those items to enhance the test validity \citep{magis2010general}.

\subsection{Psychometric alignment metric}
Assuming parameter invariance across cognitively equivalent populations allows us to measure population-level (mis-)alignment by analyzing the correlation among their parameters. Inspired by the concept of parameter invariance in IRT, we develop a metric for quantifying how well LMs align with a human population distribution of knowledge on a set of test items.

Consider a group of $N$ people $\{h_1, h_2,...,h_N\}$ and a test with a set of $M$ items $\{q_1, q_2,...,q_M\}$ with true answers $\{y_1, y_2,...,y_M\}$. We observe their responses and record them in a matrix $R_h \in \mathbb{R}^{N \times M}$, where $(R_h)_{ij}$ represents the response of the $i$-th person to the $j$-th item. To compare this with LMs, we prompt $N$ instances of LMs, labeled as ${m_1, m_2,...,m_N}$, to answer the same test items and create a similar matrix of LM responses $R_m \in \mathbb{R}^{N \times M}$.

We start by estimating the difficulty of each test item for humans $\{b_{h1}, b_{h2},...,b_{hM}\}$ by fitting an IRT model on $R_h$. Similarly, we infer the item difficulties for the LM population $\{b_{m1}, b_{m2},...,b_{mM}\}$ by applying the IRT model to $R_m$. Our evaluation metric is the Pearson correlation between these two sets of item difficulties:
\vspace{-2mm}
\begin{equation}
    \textrm{psychometric alignment} = \textrm{corr}(\{b_{h1}, b_{h2},...,b_{hM}\}, \{b_{m1}, b_{m2},...,b_{mM}\}).
    \label{eq:metric}
\end{equation}
\vspace{-6mm}

This metric intuitively assesses whether two populations perceive the relative difficulty of test items similarly. To demonstrate the robustness, sensitivity, and stability of this new metric, we conduct a series of psychometrics simulations. We start by defining 50 test items with true difficulty levels sampled from a normal distribution. We then sample two groups of students from some known ability distributions and simulate their responses to the 50 items using the 1PL model. In some conditions, we introduce group differences by swapping the true item difficulties between groups while simulating responses, thus creating items with differential item functioning (DIF). We then use the mirt package \citep{chalmers2012mirt} to estimate the item difficulties from each group's responses and calculate the correlation between these two sets of item difficulties (i.e., psychometric alignment between two groups). We repeat each simulation condition 20 times to calculate the mean and standard deviation of these correlations. Results in Table~\ref{tab: irt-simulation} highlight three characteristics of our metric: 

\textbf{Robustness}: Our metric shows high psychometric alignment when two groups of students, despite differences in size and latent abilities, perceive item difficulties similarly (no DIF items), as demonstrated in simulation conditions 1-3.
\newline
\textbf{Sensitivity}: The psychometric alignment between two groups decreases as the number of DIF items increases, even if the two groups have similar latent ability distributions (shown in simulation conditions 3-6). 
\newline
\textbf{Stability}: The psychometric alignment between two groups increases as the sizes of the groups increase, as shown in conditions 3-6 with varying $N1$ and $N2$. 

\begin{table*}
\caption{The psychometric alignment metric demonstrates several critical features: (a) Robustness, where the correlations accurately reflect true alignment, even when the two groups have varying ability distributions, as seen in conditions 1-3; (b) Sensitivity, which becomes apparent as misalignment increases with the number of DIF items—items that function differently across groups, observed in conditions 3-6; and (c) Stability, where psychometric alignment improves as group sizes increase, demonstrated in conditions 3-6. $N1$ and $N2$ represent the sizes of Group 1 and Group 2, respectively.}
\label{tab: irt-simulation}
\centering
\resizebox{\columnwidth}{!}{
\begin{tabular}{cccccccccc}
\toprule
\multicolumn{4}{l}{} & \multicolumn{1}{c}{Psychometric alignment} & \multicolumn{1}{c}{Psychometric alignment} & \multicolumn{1}{c}{Psychometric alignment} \\
\multicolumn{1}{l}{Condition} & Group 1 ability     & Group 2 ability    & DIF           & $N1=N2=200$         & $N1=200,N2=100$  
    & $N1=N2=150$\\ \midrule
1                    & N(0, 1)                  & N(-1, 0.5)              & 0        & $0.945 \pm 0.012$         & $0.920 \pm 0.014$ & -\\
2                    & N(0, 1)                  & Unif(-2,2)              & 0               & $0.946 \pm 0.015$          & $ 0.913 \pm 0.019$ &-\\
3                    & N(0, 1)                  & N(0, 1)                 & 0                       & $0.954 \pm 0.010$   & -                  & $0.905 \pm 0.039$        \\
4                    & N(0, 1)                  & N(0, 1)                 & 5            & $0.873 \pm 0.068 $           &-& $0.863 \pm 0.044$        \\
5                    & N(0, 1)                  & N(0, 1)                 & 10          & $0.814 \pm 0.058$           &- & $0.784 \pm 0.078$        \\
6                    & N(0, 1)                  & N(0, 1)                 & 50         & $0.372 \pm 0.144$              &- & $0.288 \pm 0.145$      \\
\bottomrule
\end{tabular}
\vspace{-10pt}
}
\end{table*}
\vspace{-10pt}

%% file: sections/datasets.tex
\section{Datasets}
\label{sec:dataset}

To assess how well LMs capture the human population distribution using psychometric alignment, we need datasets of human responses to a set of test items. It is important that we have the full text content of the items in order to enable LM evaluation; this content is missing from most available educational datasets. We now describe three real-world datasets with the required information. 

\begin{figure}
\begin{minipage}{0.4\textwidth}
\begin{figure}[H]
    \centering   \includegraphics[width=\columnwidth]{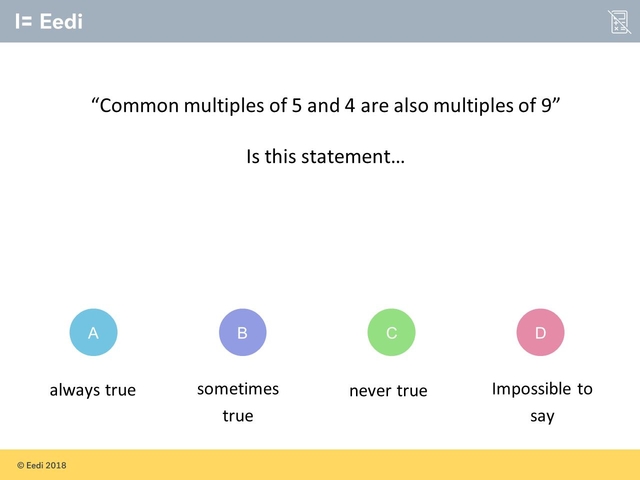}
    \caption{An example of a question from the \eedi{} dataset.}
    \label{fig:example_question}
\end{figure}
\end{minipage}
\hfill
\begin{minipage}{0.55\textwidth}
\begin{figure}[H]
    \centering   \includegraphics[width=0.55\columnwidth]{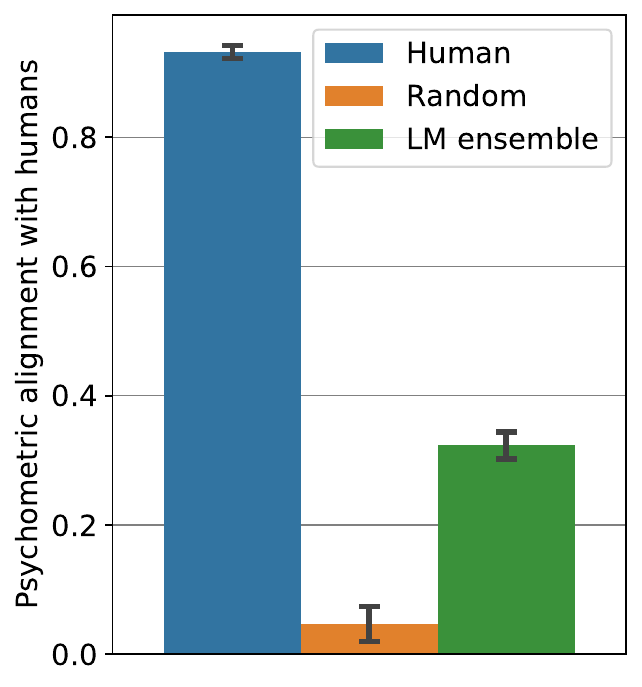}
    \caption{Ensembling different LMs does not generate an LM population that captures the distribution of knowledge in human population from the \eedi{} dataset. The error bars indicate the standard deviation.}
    \label{fig:naive_emsemble}
\end{figure}
\end{minipage}
\vspace{-12pt}
\end{figure}

\textbf{\eedi{}: Math diagnostic assessments}$\quad$ The \eedi{} dataset is built on the NeurIPS 2020 Education
Challenge dataset \citep{wang2020instructions}, provided by the Eedi online educational platform\footnote{\url{https://eedi.com}}. It contains student responses to math multiple-choice questions (see Figure \ref{fig:example_question}) collected between September 2018 and May 2020. The NeurIPS 2020 Education
Challenge dataset provided question content in image format (e.g., Figure \ref{fig:example_question}) without accompanying texts. With permission from Eedi, we have extracted the text from these question images and released this modified dataset. We excluded questions with graphs or diagrams since most current language models do not support visual inputs. The modified dataset contains $573$ unique questions and $443,433$ responses to these questions from $2,287$ students, along with data on their age (mostly 11-12 years), gender, and socioeconomic status. For our analysis, we randomly selected a test set of 50 questions and 150 students who answered all these questions.\footnote{We need at least 150 students to get stable IRT parameters in the \eedi{} domain. See Figure \ref{fig:eedi_finetune_1pl}.} We treat this test set as representative of the human population. All LM evaluations were conducted on the test set.

\textbf{\wordbank{}: Vocabulary development}$\quad$ The \wordbank{} dataset is from the WordBank database\footnote{\url{github.com/langcog/wordbankr}} \citep{frank2017wordbank}. We focus on the English (American) subset, which includes responses from $5,520$ children aged between 16 and 30 months. Each child responded to 680 vocabulary items. We only consider items that are words. The responses, reported by parents, are binary and indicate whether the child can produce each word. The dataset also contains demographic details for each child such as age, gender, ethnicity, and the education level of the mother. We randomly selected a test set of 50 words and 150 children to represent the human population. 

\textbf{\duolingo{}: App-based language learning}$\quad$ The \duolingo{} dataset is from the 2018 Duolingo Shared Task on Second Language Acquisition Modeling (SLAM)\footnote{\url{sharedtask.duolingo.com/2018.html}} \citep{settles2018second}. This dataset contains anonymized data from users of the educational application Duolingo\footnote{\url{duolingo.com}}. We focus on the subset of English speakers learning Spanish through lesson sessions. Each user's data consists of a series of binary responses to vocabulary words, with each word presented multiple times. Following the approach described in \cite{wu2020variational}, we adapted this dataset for Item Response Theory modeling by averaging responses to each vocabulary item, rounding the average score to a binary outcome (0 or 1). For instance, if a user was shown the word ``hola" 10 times and correctly translated the word 5 times, the average score would be 0.5 and rounded to 1. After processing, the dataset includes $2,783$ vocabulary words and $573,321$ responses from $2,640$ users, with missing data due to user dropout. The dataset also includes additional user information such as country and device type. We randomly selected a test set of 50 words and 500 users who have responded to these words. In this domain, we needed a larger sample to get stable IRT parameters compared to the other domains because the dataset is more sparse (with more positive labels than negative labels). The percentage of correct responses is over $87\%$.

\section{The importance of psychometric alignment}
\label{sec:the_importance_of_psychometric_alignment}
We first illustrate the importance of our psychometric alignment metric (Eq.~\ref{eq:metric}) using the \eedi{} dataset as an example. While it might seem straightforward to compare populations based on summary statistics such as person accuracies or test scores \citep{xu2023leveraging, zelikman2023generating, chenevaluating}, these metrics do not capture the distribution of knowledge across individual test items and can be misleading when assessing the similarity of two populations and the significance of specific items. To illustrate this, we modified the \eedi{} dataset to create a synthetic population by randomly shuffling responses.

\begin{figure}
\begin{minipage}{0.68\textwidth}
\begin{figure}[H]
    \begin{subfigure}[b]{\textwidth}
        \centering
        \includegraphics[width=0.7\textwidth]{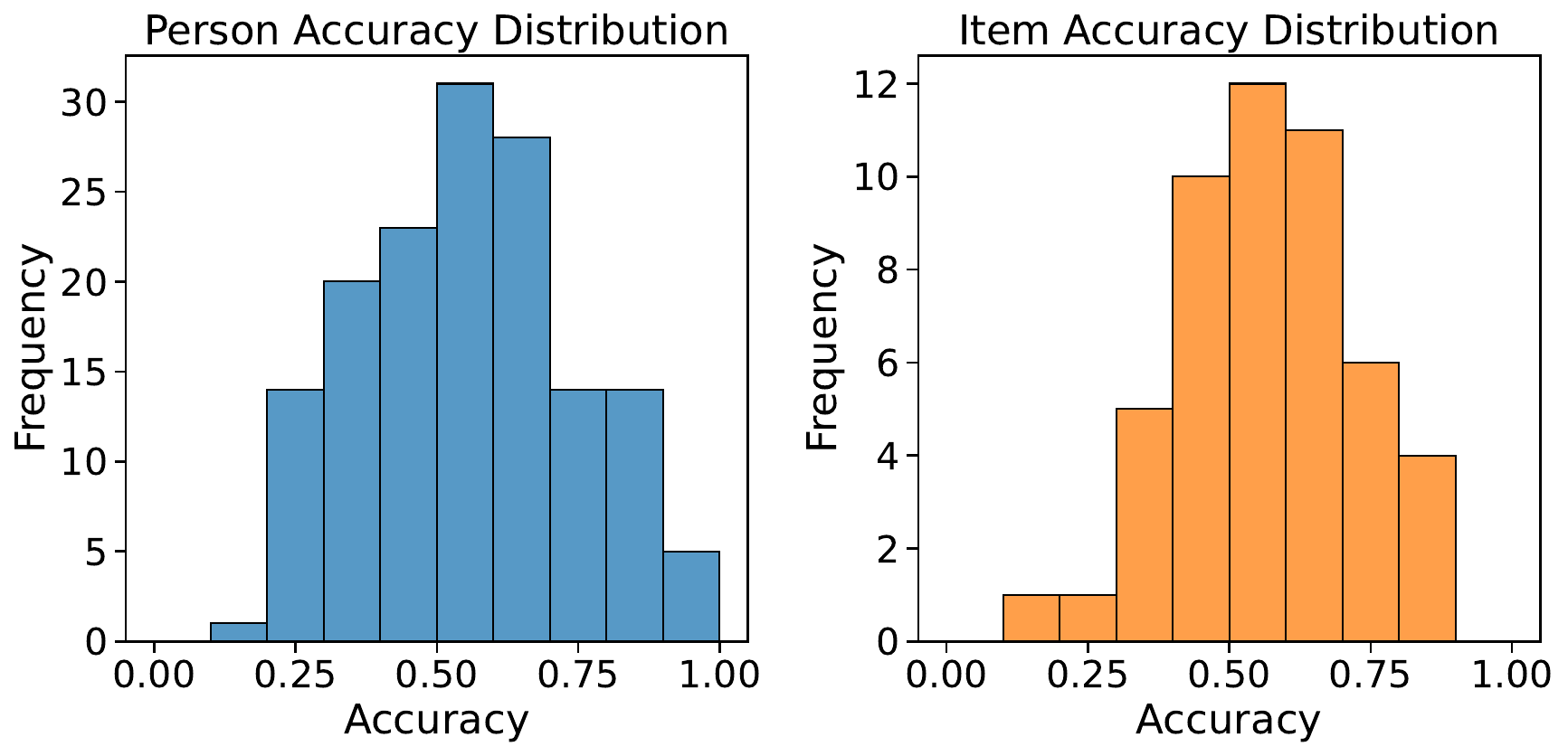}
        \caption{Human}
        \label{fig:human_dist}
    \end{subfigure}
    \enskip 
    \begin{subfigure}[b]{\textwidth}
        \centering
        \includegraphics[width=0.7\textwidth]{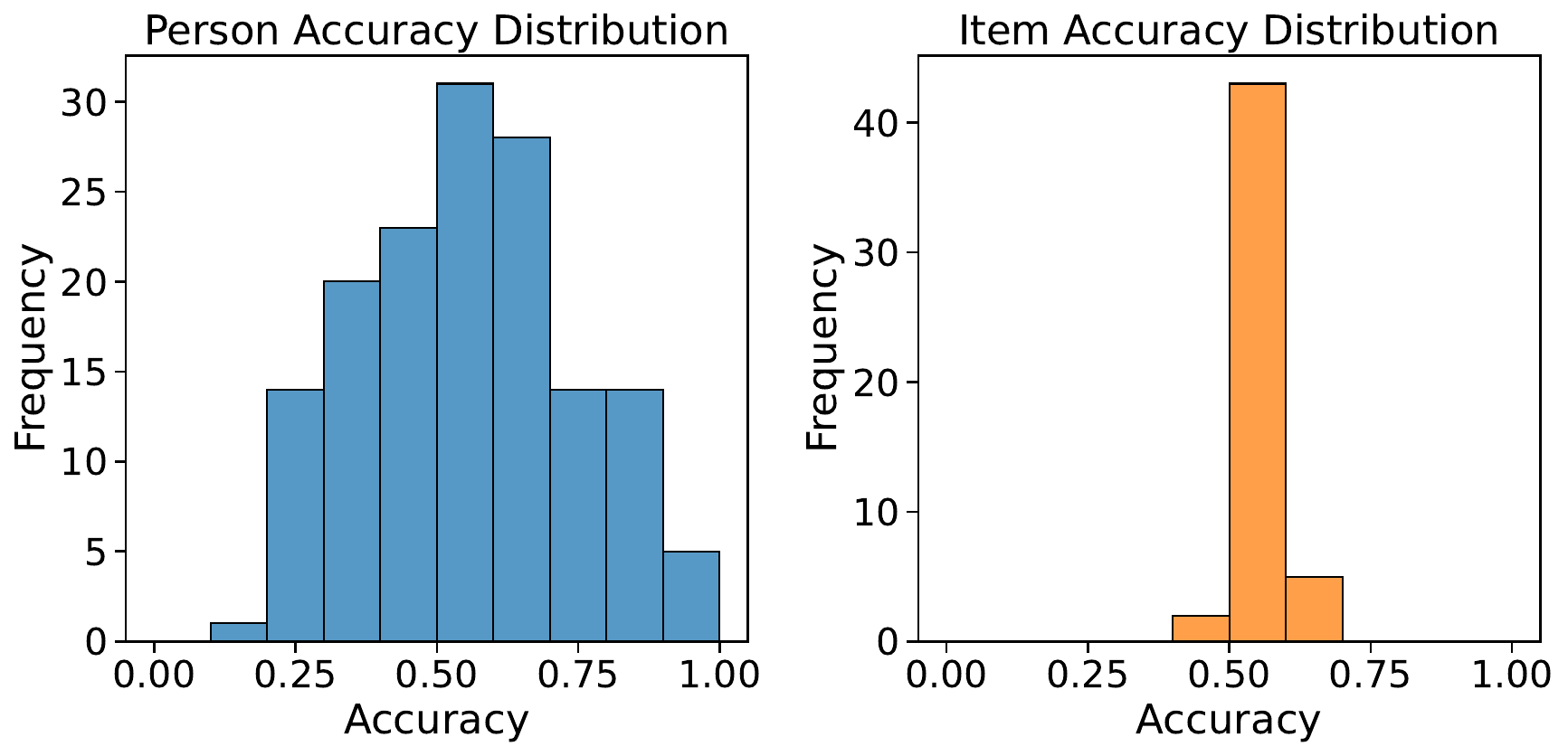}
        \caption{Synthetic}      \label{fig:synthetic_match_person_dist}
    \end{subfigure}
    \caption{\ref{fig:human_dist} shows the person accuracy distribution and item accuracy distribution of the \eedi{} data. We generate a synthetic population by randomly shuffling responses within each person (\ref{fig:synthetic_match_person_dist}). }
    \label{fig:human_vs_synthetic_dist}
\end{figure}
\end{minipage}
\hfill
\begin{minipage}{0.3\textwidth}
\begin{figure}[H]
    \centering   \includegraphics[width=0.8\columnwidth]{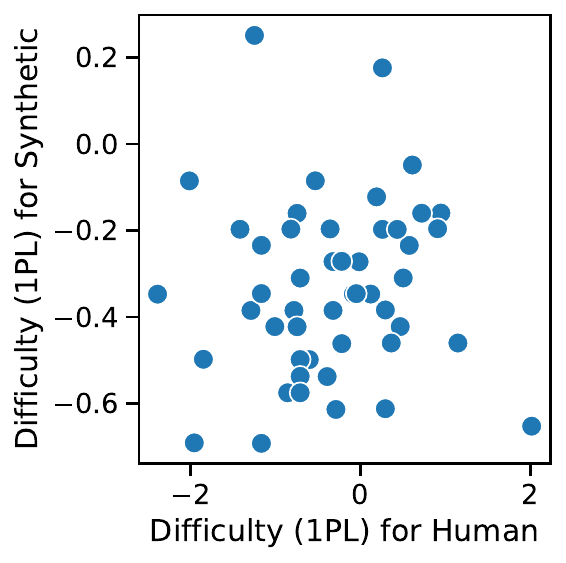}
    \caption{Some items that are easy for humans are hard for the Synthetic population, indicating that even when two populations show similar overall score distributions, they might possess distinct latent abilities and respond differently to the same questions. There is no significant correlation between the difficulty (1PL) parameters of the two populations (Pearson $r=0.07, p > 0.05$).}
    \label{fig:human_vs_synthetic_diff1pl}
\end{figure}
\end{minipage}
\vspace{-12pt}
\end{figure}

Figure \ref{fig:human_dist} shows the distribution of person accuracy/test score (i.e., the percentage of items that each student answers correctly) and the distribution of item accuracy (i.e., the percentage of students who answer each item correctly) in the original data. Figure \ref{fig:synthetic_match_person_dist} shows these distributions for a synthetic population, which mirrors the human population in terms of overall score distribution but shows variation in item accuracy. We generated this synthetic population by randomly shuffling the original human responses within each person. Although the synthetic population aligns perfectly with the human population in terms of overall test score distribution, the two differ significantly in their assessment of which items are difficult. Figure \ref{fig:human_vs_synthetic_diff1pl} shows the item difficulties estimated using the 1PL IRT model for each population, with higher values indicating greater difficulty. This suggests that populations with similar score distributions can exhibit markedly different item functionalities—an important consideration if we intend to use the synthetic population data to develop new test items or educational materials for humans.

%% file: sections/4_prompting.tex
\section{Prompting-based ensemble}
We use our psychometric alignment metric to evaluate existing LMs on three datasets: \eedi{}, \wordbank{}, and \duolingo{}. We start by assessing the default psychometric alignment of an ensemble of LMs without prompting them to mimic any specific group. Then, we explore the impact of various group-specific prompting strategies on psychometric alignment.

\subsection{Control conditions}

In the ensuing evaluations we compare to two control conditions:
\vspace{-3mm}
\paragraph{Human (positive control):} For the \eedi{} and \wordbank{} datasets, we construct 20 datasets where each dataset consists of 150 students randomly selected from the corresponding data but not in the test set. For the \duolingo{} dataset, we similarly construct 20 datasets, each consisting of 500 students. In all domains, we only consider students who have complete responses to all 50 items that are in the test set. We then calculate the psychometric alignment metric by comparing the difficulty parameters derived from the test set against those from each of the 20 human datasets. This allows us to assess whether the item parameters are consistent within the human population and represents an estimate of ceiling psychometric alignment.
\vspace{-3mm}
\paragraph{Random (negative control):} For \eedi{}, we construct a synthetic response matrix where each user has a $25\%$ chance of answering each question correctly, aligning with the probability of guessing correctly in a four-option multiple-choice format. For \wordbank{} and \duolingo{}, where we only have binary labels, we generate a response matrix where each user has a $50\%$ chance of getting each item correct. We calculate the psychometric alignment metric by comparing the difficulty parameters obtained from the random response matrix with those from each human sample. This represents floor psychometric alignment.

\subsection{Ensembling different LMs}
To simulate a human population, we need to create a population of LMs. We start by exploring whether an ensemble of different LMs can capture the response variations in a human population. To do so, we evaluate 10 open-source LMs of varying capabilities (see details in Appendix \ref{appendix:prompts_for_lm_ensemble}) on the \eedi{} dataset and mix responses from these LMs to create an \textbf{LM-ensemble} response matrix. We prompt each LM to answer each of the 50 questions from the held-out test set (see Section \ref{sec:dataset}) 15 times. We use a zero-shot or a few-shot prompt depending on the LM's capability (see Appendix \ref{appendix:prompts_for_lm_ensemble}). We also vary the temperature settings (0, 0.7, and 1) to diversify responses, resulting in a total of 150 sets of responses to the 50 questions. We then fit the 1PL IRT model on this response data to estimate the item difficulty parameters. We selected the \eedi{} dataset because, unlike vocabulary or language acquisition tasks, the mathematical capabilities of LMs continue to show significant variations across LMs and datasets, which is crucial for both simulating human variations and fitting IRT models effectively.

Figure \ref{fig:naive_emsemble} shows the mean and standard deviation of the psychometric alignment with humans for \textbf{LM-ensemble} and controls. We observe high stability in the item difficulty parameters across human samples of size 150, with a correlation exceeding $0.9$. However, neither the \textbf{Random} nor the \textbf{LM-ensemble} methods achieve a strong alignment with the human population, indicating discrepancies in LM responses compared to human variability. This highlights challenges in achieving psychometric alignment by merely combining existing LMs of varying sizes or capabilities.

\subsection{Persona-based prompting}

\begin{table*}
\caption{The mean and standard deviation of the psychometric alignment of LMs when prompted using different persona-based prompts. ``Invalid'' indicates we cannot fit an IRT model on the responses due to lack of variability.}
\label{table:prompting}
\centering
\resizebox{\columnwidth}{!}{
\begin{tabular}{llllllllll}
\toprule
\multicolumn{2}{l}{Dataset \& prompt} & Human & Random & Mistral-7b & Llama-8b & Llama-70b & GPT-3.5 & GPT-4 \\ \midrule
\multirow{4}{*}{\eedi} & \persona & - & - & $0.378 \pm 0.03$ & $0.503 \pm 0.04$ & $0.314 \pm 0.02$ & $0.309 \pm 0.03$ & -  \\
& \personacot & - & - & $0.403 \pm 0.03$ & $0.567 \pm 0.04$ & $0.405 \pm 0.03$ & $0.331 \pm 0.03$ & -  \\
 & \personacots & - & - & $\bf 0.448 \pm 0.03$ & $\bf 0.582 \pm 0.04$ & $\bf 0.436 \pm 0.02$ & $\bf 0.582 \pm 0.02$ & $0.202 \pm 0.04$  \\
 & N/A & $0.932 \pm 0.01$ & $0.047 \pm 0.03$ & - & - & - & - & -  \\ \midrule
\multirow{4}{*}{\wordbank} & \persona & - & - & $0.221 \pm 0.02$ & $0.400 \pm 0.02$ & $\bf 0.608 \pm 0.01$ & Invalid & -  \\
& \personacot & - & - & $\bf 0.351 \pm 0.02$ & $\bf 0.404 \pm 0.02$ & $0.443 \pm 0.02$ & $0.153 \pm 0.02$ & -  \\
 & \personacots & - & - & $0.267 \pm 0.02$ & $0.256 \pm 0.02$ & $0.563 \pm 0.02$ & $\bf 0.300 \pm 0.02$ & $0.302 \pm 0.02$  \\
 & N/A & $0.974 \pm 0.00$ & $-0.094 \pm 0.02$ & - & - & - & - & -  \\  \midrule
\multirow{4}{*}{\duolingo} & \persona & - & - & $\bf 0.321 \pm 0.06$ & $\bf 0.427 \pm 0.07$ & $0.188 \pm 0.08$ & $0.287 \pm 0.07$ & -  \\
& \personacot & - & - & $0.309 \pm 0.08$ & $0.337 \pm 0.10$ & $0.234 \pm 0.10$ & $\bf 0.313 \pm 0.08$ & $0.271 \pm 0.08$ \\
 & \personacots & - & - & $0.236 \pm 0.08$ & $0.427 \pm 0.08$ & $\bf 0.276 \pm 0.10$ & $0.173 \pm 0.11$ & -  \\
 & N/A & $0.741 \pm 0.09$ & $0.067 \pm 0.05$ & - & - & - & - & -  \\  \bottomrule
\end{tabular}
}
\end{table*}

Recent papers have demonstrated that LMs can more accurately capture certain behaviors of a human group (e.g., voting preferences) when prompted with group-specific demographic information in their context \citep{argyle2023out, santurkar2023whose}. We refer to this approach as persona-based prompting and explore whether asking an LM to pretend to be individuals with different personas can steer the model to better represent the human population. For example, before asking the LM to respond to a problem we can create personas such as "Pretend that you are an 11-year-old student. Your gender is female. You are eligible for free school meals due to
being financially disadvantaged."

We explore three prompting strategies that use persona descriptions like the above:
\begin{enumerate}
    \item \persona: We ask the LM to solve the problem given the persona.
    \item \personacot{} (persona + CoT): We ask the LM to reason about its ability to solve the problem given the persona before providing an answer. This is inspired by the Chain-of-Thought prompting method \citep{wei2022chain} that asks an LM to present explicit intermediate reasoning steps to further enhance its own reasoning capability.
    \item \personacots{} (persona + CoT + structure): We ask the LM to reason about its ability to solve the problem given the persona and explicitly structure its response based on this assessment.
\end{enumerate}
For examples of these prompts and personas, see Figure \ref{fig:persona_prompt_example} in Appendix. We evaluate 5 popular LMs using each of these prompting methods across our three datasets. In each evaluation, we use real student demographic information from the test set to construct each student's persona and prompt the LM to generate responses within the context of prior questions and responses associated with the same student. We only evaluated GPT-4 in a few conditions because it is expensive.

Table \ref{table:prompting} shows evaluation results on the three datasets: \eedi{}, \wordbank{}, and \duolingo{}. We set temperature=0.7 for all LMs. Temperature 1 yields similar results (see Table \ref{table:prompting_temp_1} in Appendix). As before, there is a large gap between human and random controls. For \eedi{}, \personacots{} is the most effective prompt for all models. However, the best prompt for \wordbank{} and \duolingo{} varies depending on the model used. Interestingly, larger LMs tend to show poorer psychometric alignment than smaller ones. For instance, Llama-8b either matches or exceeds the performance of both GPT-3.5 and GPT-4 in all three domains and outperforms Llama-70b in two of them. This suggests that steering larger LMs to represent specific groups through persona-based prompting may be more challenging, despite their enhanced capability to follow prompt instructions. We also experimented with varying the number of features/attributes used in the personas and found that adding more features in the persona did not increase the psychometric alignment further (see Appendix \ref{appendix:prompting_ablations}), suggesting the need for alternate approaches to improve psychometric alignment.

%% file: sections/finetuning.tex
\section{Fine-tuning LMs on student response data}
\label{sec:finetuning}
\begin{figure}[ht]
    \centering
    \includegraphics[width=\linewidth]{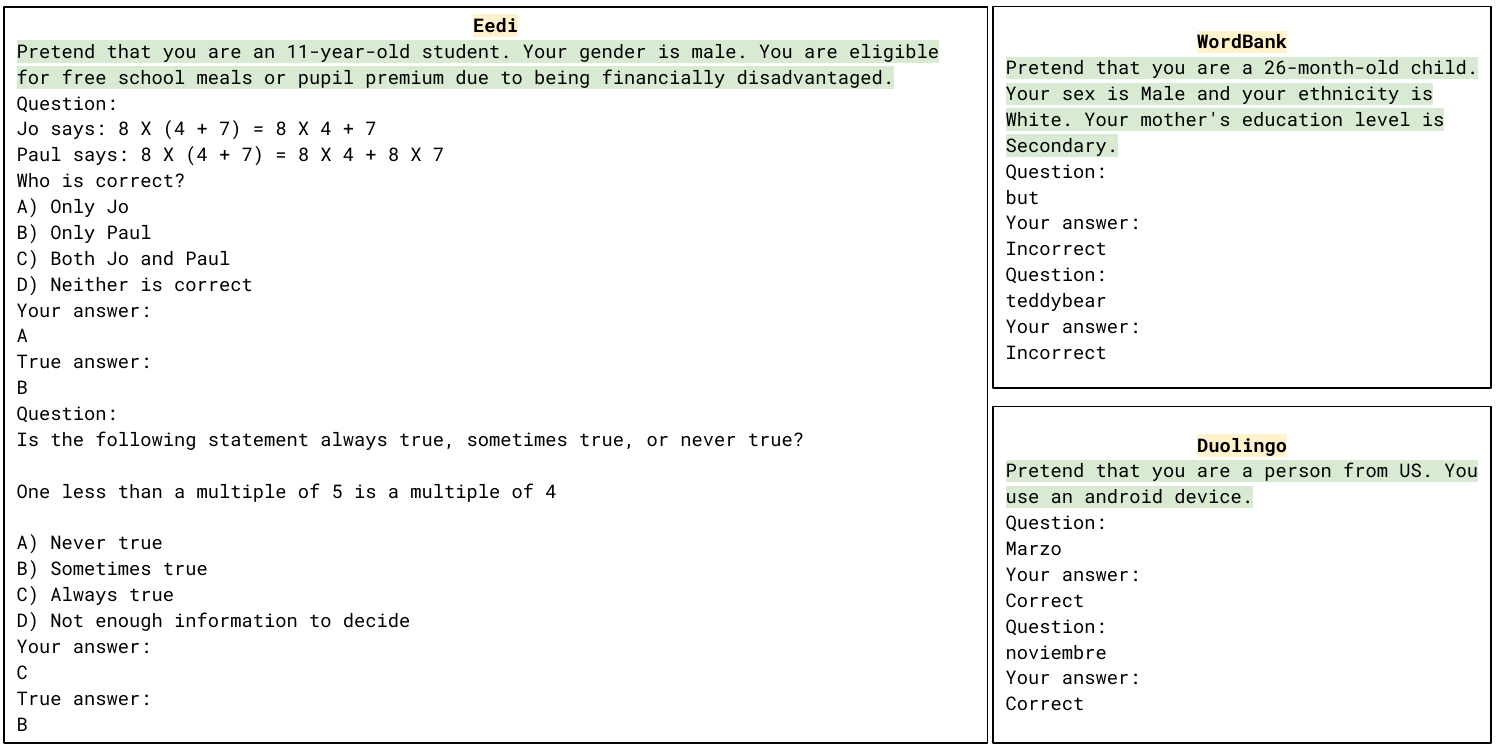}
    \caption{Example training data.}
    \label{fig:finetune_data_example}
\end{figure}

In educational contexts, researchers have considered fine-tuning an LM on student response data to create student simulators for generating or evaluating test items \citep{srivastava2021question, zelikman2023generating}. Therefore, we explore if fine-tuning LMs on student response data can enhance the psychometric alignment between LMs and humans on \emph{unseen} test items. We train three different LMs (Mistral-7b, Llama-8b, and Deepseek-7b) to predict student responses from their attributes (persona) and historical data. Each training example consists of a sampled student's persona and a randomly-selected subset of that student's item-response pairs (see examples in Figure \ref{fig:finetune_data_example}). For \duolingo{} and \wordbank{}, since we only have binary labels, responses are classified as either ``Correct'' or ``Incorrect.'' For \eedi{}, which collects actual student responses (e.g., selected letters), we include both the student's chosen answer and the true answer. To fine-tune LMs, we use Low-Rank Adaptation (LoRA) \citep{hu2021lora} with an adaptor rank of $r=32$ and $lora\_alpha=64$. We train the LMs using different amounts of student data, by varying the number of unique students included. The training data do not include any students or items that are in the test set. Further details on the training data and hyperparameters are available in Appendix~\ref{appendix:finetuning}.

For evaluation, we ask the LM to simulate a student's response to each item in the test set based on the student's persona, prior items, and LM-predicted responses. For \eedi{}, we also include the true answer to each prior item. The evaluation prompts follow the same template used for training (see Figure \ref{fig:finetune_data_example}).

\begin{figure}[t]
    \begin{subfigure}[b]{0.32\textwidth}
        \includegraphics[width=\textwidth]{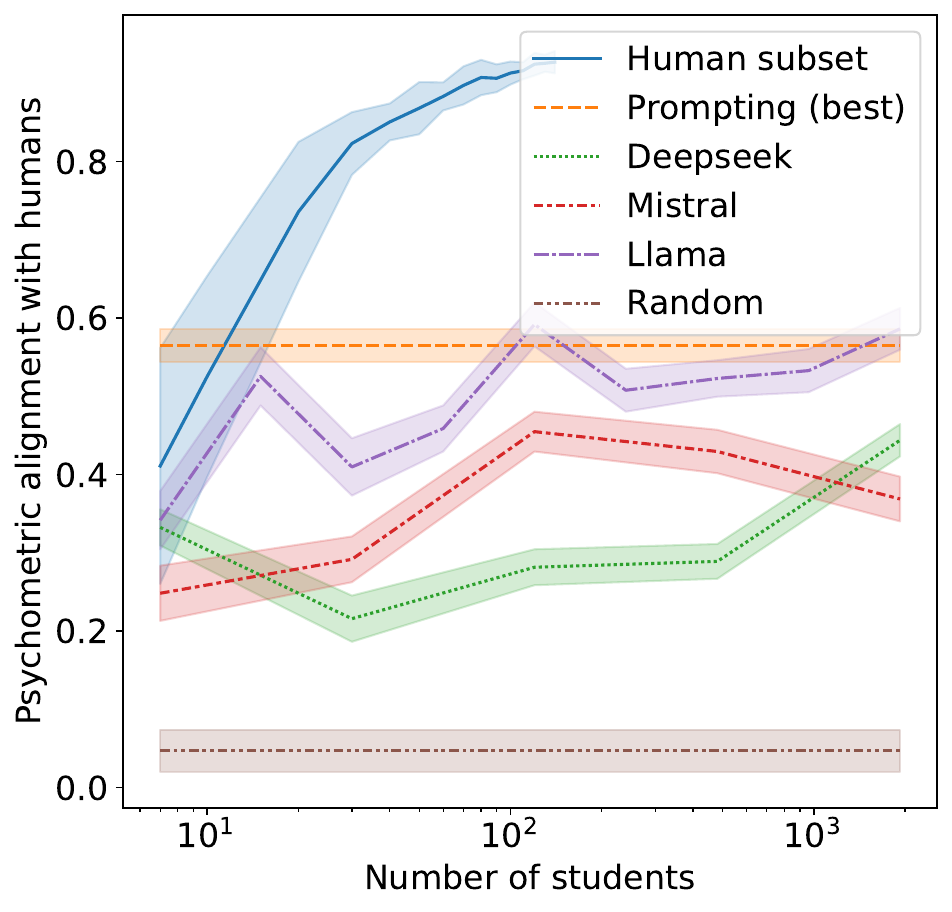}
        \caption{\eedi{}}
        \label{fig:eedi_finetune_1pl}
    \end{subfigure}
    \enskip 
    \begin{subfigure}[b]{0.32\textwidth}
        \includegraphics[width=\textwidth]{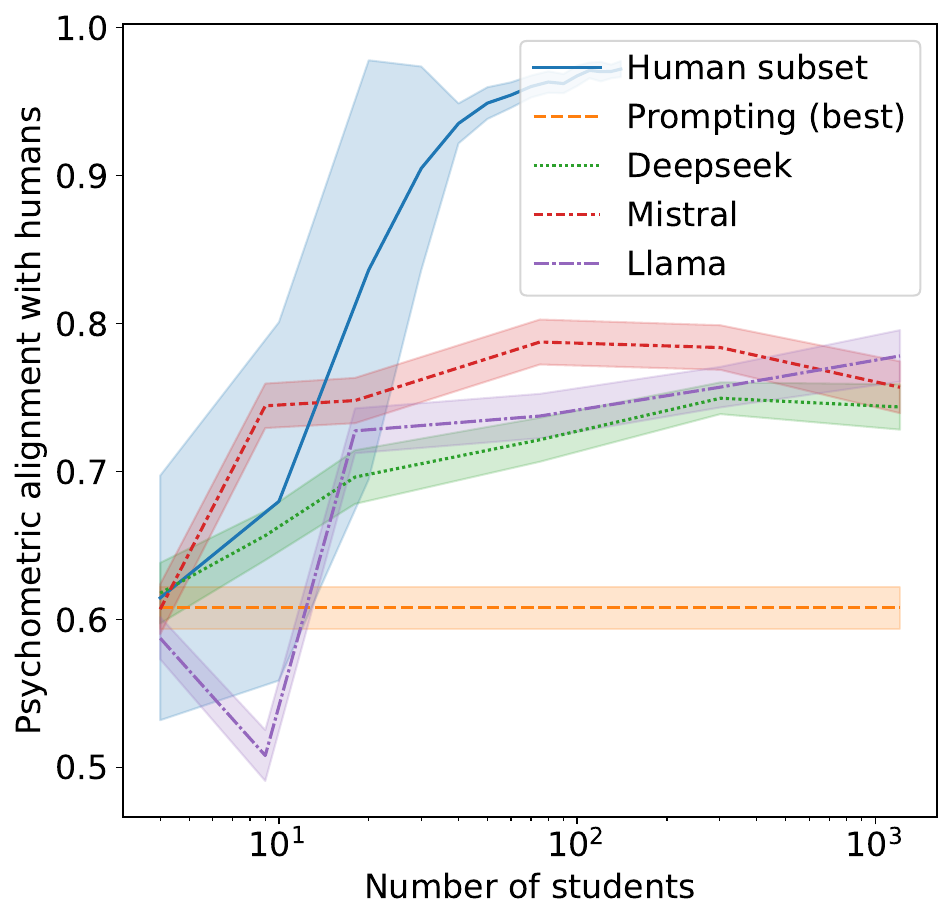}
        \caption{\wordbank{}}
        \label{fig:wordbank_finetune_1pl}
    \end{subfigure}
    \enskip 
    \begin{subfigure}[b]{0.32\textwidth}
        \includegraphics[width=\textwidth]{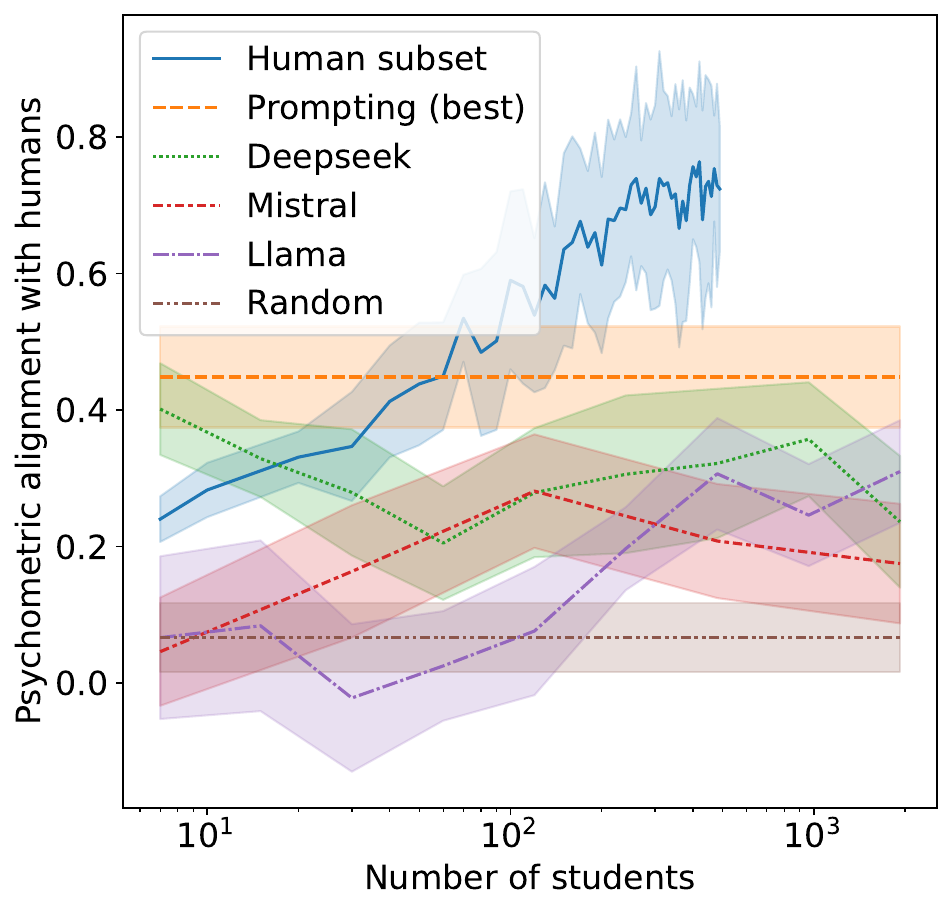}
        \caption{\duolingo{}}
        \label{fig:duolingo_finetune_1pl}
    \end{subfigure}
    \\ 
    \caption{Fine-tuned LMs outperform the best prompting baseline on \wordbank{}, but not in the other domains. Error bars indicate the standard deviation.}
\end{figure}

We find that fine-tuning these base models on student response data does not improve over the best prompting baseline (see Table \ref{table:prompting}) for \eedi{} and \duolingo{} (see Figure \ref{fig:eedi_finetune_1pl} and Figure \ref{fig:duolingo_finetune_1pl}). On \eedi{}, Llama-8b outperforms the other LMs, but there is no significant difference across the LMs for \duolingo{}. However, fine-tuned LMs outperform the best prompting baseline for \wordbank{} (Figure \ref{fig:wordbank_finetune_1pl}), using historical data from just a few students. This improvement could be due to higher similarities between the training and test set items. We also compare all LMs to the human baseline that uses real human data on the test set items (referred to as ``human subset''); however, we do not expect any of the other methods to match the ``human subset'' performance as they do not have access to the response data on the test set items. The ``human subset'' baseline helps gauge how much real data could potentially be saved by training LMs on historical data. For instance, the top-performing fine-tuned LM for \wordbank{} reaches the psychometric alignment that could be achieved with 10-20 real students' data on the test set items.

\vspace{-10pt}

%% file: sections/conclusion.tex
\section{Limitations}

There are several limitations. First, we focus on the 1PL IRT model because it is widely used and fits the \eedi{} dataset best, but further insights might be gained by examining more sophisticated IRT models such as those considering multiple latent ability dimensions. Second, we acknowledge that no dataset can fully represent the entire human population. For example, our \eedi{} dataset is limited to students in England who choose to use the platform. Third, the datasets we use were not collected in typical assessment settings and may violate certain IRT assumptions (e.g., no learning between individual responses).

\section{Conclusion}

We propose an evaluation metric to assess the extent to which LMs capture the distribution of human knowledge. We demonstrate that our metric is more robust than traditional ones. We view our metric as a tool to enable people to better understand LM behaviors and identify potential representation failures when using LMs to simulate humans. 

%% file: sections/appendix.tex
\section{Appendix / supplemental material}

\subsection{LM ensembling implementation details}
To create the \textbf{LM-ensemble}, we consider Mistral-7B-v0.1, llemma\_7b, llemma\_34b, deepseek-math-7b-base, deepseek-math-7b-instruct, deepseek-math-7b-rl, Meta-Llama-3-8B, Meta-Llama-3-8B-Instruct, Meta-Llama-3-70B, and Meta-Llama-3-70B-Instruct.

\begin{figure}[H]
    \centering
    \includegraphics[width=0.6\linewidth]{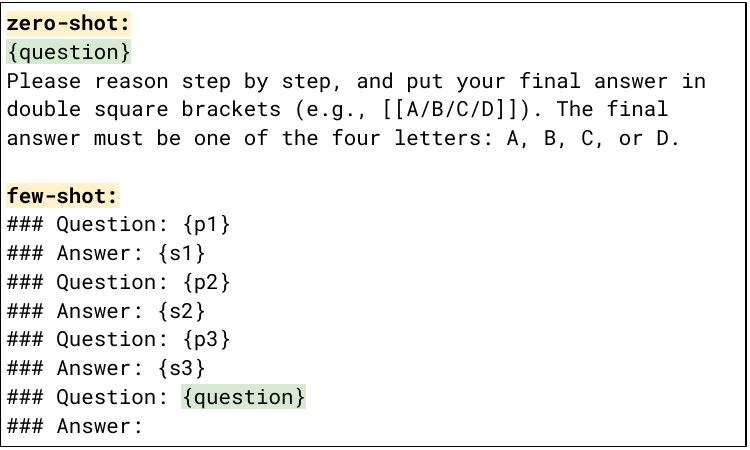}
    \caption{The few-shot prompt has three example question-answer pairs.}
    \label{fig:lm_ensemble_prompt}
\end{figure}

For the base LMs (Mistral-7B-v0.1, llemma\_7b, llemma\_34b, deepseek-math-7b-base, Meta-Llama-3-8B, and Meta-Llama-3-70B), we need to use a few-shot prompt (see Figure \ref{fig:lm_ensemble_prompt}) to ensure their responses to test items are in a consistent format. For the other instruction-tuned LMs, using a zero-shot prompt is sufficient.

\label{appendix:prompts_for_lm_ensemble}

\subsection{Persona-based prompting examples}
\begin{figure}[H]
    \centering
    \includegraphics[width=0.9\linewidth]{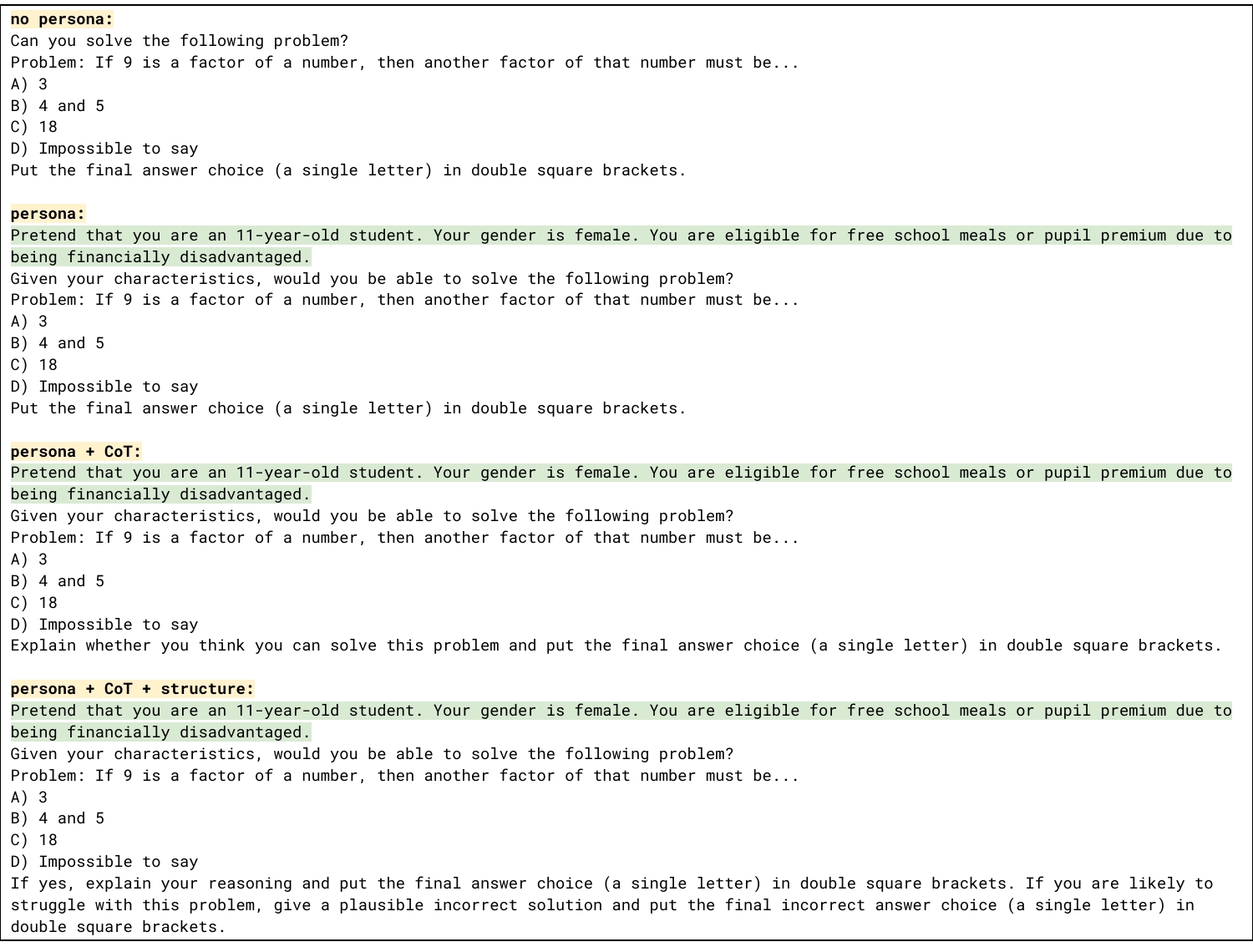}
    \caption{Example input prompt for each prompting method. The persona is highlighted in green.}
    \label{fig:persona_prompt_example}
\end{figure}

\subsection{Prompting ablations}
\label{appendix:prompting_ablations}

The \eedi{} dataset contains three attributes (see Section \ref{sec:dataset}). We generate 5 additional math-relevant features: numerical proficiency, working memory, math anxiety, math importance, parental involvement.

\begin{figure}[H]
    \centering
    \includegraphics[width=0.35\linewidth]{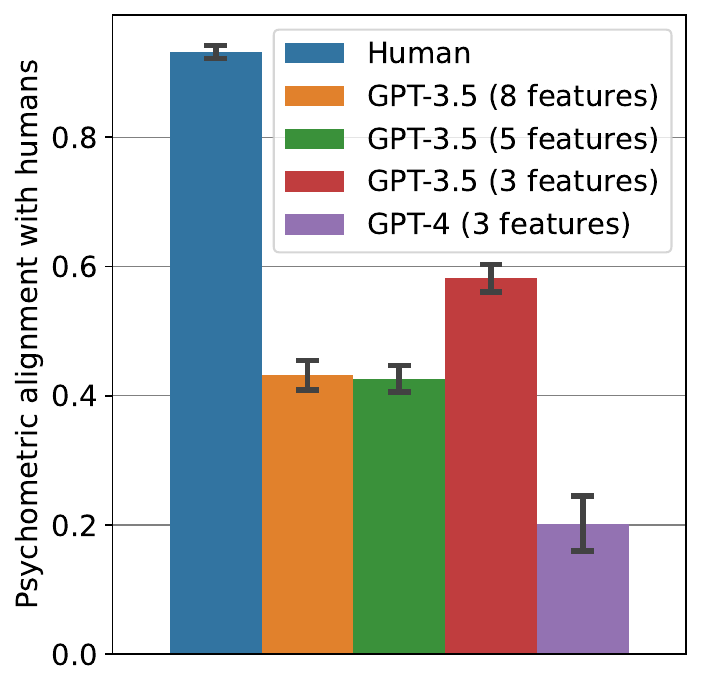}
    \caption{Adding more features in personas does not achieve higher psychometric alignment. GPT-4 is less human-aligned compared to GPT-3.5. All LMs use a temperature of 0.7. We use \personacots{} prompting for all LMs. Error bars indicate standard deviation.}
    \label{fig:gpt_prompting_num_attributes}
\end{figure}

\begin{table*}[h]
\caption{The mean and standard deviation of the psychometric alignment of LMs when prompted using different persona-based prompts. All LMs use a temperature of 1. 
}
\label{table:prompting_temp_1}
\centering
\resizebox{\columnwidth}{!}{
\begin{tabular}{llllllllll}
\toprule
\multicolumn{2}{l}{Dataset \& prompt} & Human & Random & Mistral-7b & Llama-8b & Llama-70b & GPT-3.5  \\ \midrule
\multirow{4}{*}{\eedi} & \persona & - & - & $0.448 \pm 0.03$ & $0.510 \pm 0.04$ & $0.319 \pm 0.04$ & $0.363 \pm 0.07$   \\
                    & \personacot & - & - & $0.441 \pm 0.03$ & $0.525 \pm 0.04$ & $0.415 \pm 0.03$ & $0.353 \pm 0.03$  \\
                   & \personacots & - & - & $\bf 0.464 \pm 0.03$ & $\bf 0.537 \pm 0.04$ & $\bf 0.510 \pm 0.05$ & $\bf 0.476 \pm 0.02$  \\
                       & N/A & $0.932 \pm 0.01$ & $0.047 \pm 0.03$ & - & - & - & -  \\ \midrule
\multirow{4}{*}{\wordbank} & \persona & - & - & $\bf 0.363 \pm 0.02$ & $0.373 \pm 0.02$ & $\bf 0.630 \pm 0.02$ &  \\
                        & \personacot & - & - & $0.317 \pm 0.03$ & $\bf 0.398 \pm 0.02$ & $0.548 \pm 0.02$ & $0.219 \pm 0.02$  \\
                       & \personacots & - & - & $0.230 \pm 0.02$ & $0.331 \pm 0.02$ & $0.594 \pm 0.02$ & $\bf 0.290 \pm 0.02$ \\
                                & N/A & $0.974 \pm 0.00$ & $-0.094 \pm 0.02$ & - & - & - & -   \\  \midrule
\multirow{4}{*}{\duolingo} & \persona & - & - & $ 0.327 \pm 0.07$ & $ 0.406 \pm 0.08$ & - & $\bf 0.309 \pm 0.07$   \\
                        & \personacot & - & - & $\bf 0.343 \pm 0.09$ & $0.415 \pm 0.09$ & $0.227 \pm 0.10$ & $0.295 \pm 0.08$   \\
                       & \personacots & - & - & $0.220 \pm 0.09$ & $\bf 0.450 \pm 0.09$ & $\bf 0.291 \pm 0.10$ & $0.176 \pm 0.11$   \\
 & N/A & $0.741 \pm 0.09$ & $0.067 \pm 0.05$ & - & - & - & -   \\  \bottomrule
\end{tabular}
}
\end{table*}

\subsection{Fine-tune on student data}
\label{appendix:finetuning}

\textbf{Training data}\quad For each dataset, we randomly split the dataset into training and validation set by the user. Specifically, we use $10\%$ of the users as the validation set. We remove all questions in the test set from the training and validation data. To create each data point for the \eedi{} domain, we randomly sample 4-11 question-response pairs from each user's quiz sequence and repeat 20 times for each user's quiz. For \duolingo{} and \wordbank{}, since the items are shorter, we randomly sample up to 50 question-response pairs from each user's data and repeat 200 times for each user in \duolingo{} and 100 times for each user in \wordbank{}. All question-response pairs are arranged in random order in each data point.

\textbf{LMs}\quad We use the base models of three LM classes: Mistral-7b, Llama-8b, and Deepseek-7b. Specifically, we use Mistral-7B-v0.1, Meta-Llama-3-8B, and deepseek-math-7b-base. 

\textbf{Hyperparameters}\quad To fine-tune LMs, we use Low-Rank Adaptation (LoRA) \citep{hu2021lora} with an adaptor rank of $r=32$ and $lora\_alpha=64$. All models are run in 4-bit quantization. We use the 8-bit AdamW with a learning rate fixed at 2.5e-5, and the models are trained with a batch size of 32. We set $gradient\_accumulation\_steps=1$. We selected hyperparameters based on early experiments with Mistral-7B-v0.1. We compared $r=16$ with $r=32$ and did not find a significant difference, so we consistently used $r=32$ and $lora\_alpha=64$ for all LMs. Each model is trained until there is no reduction in evaluation loss on the validation set across three successive iterations.

\textbf{Compute} \quad Each LM was fine-tuned on a single 80GB A100 GPU, with training times ranging from 1 to 48 hours depending on the size of the training data.